\documentclass[11pt,a4paper]{article}
\usepackage[hyperref]{emnlp2020}
\usepackage{times}
\usepackage{latexsym}

\usepackage{booktabs,multirow,array}
\usepackage{color}
\usepackage{graphicx} 
\usepackage{amsfonts,amsmath}
\usepackage{ulem}

\usepackage{microtype}

\aclfinalcopy % Uncomment this line for the final submission
\title{Multi-hop Inference for Question-driven Summarization\thanks{\hspace{2mm}The work described in this paper is substantially supported by a grant from the Research Grant Council of the Hong Kong Special Administrative Region, China (Project Codes: 14200719).}}

\author{Yang Deng, Wenxuan Zhang, Wai Lam \\
  The Chinese University of Hong Kong \\
  \texttt{\{ydeng,wxzhang,wlam\}@se.cuhk.edu.hk} \\}

\date{}

\begin{document}
\maketitle
\begin{abstract}
Question-driven summarization has been recently studied as an effective approach to summarizing the source document to produce concise but informative answers for non-factoid questions. 
In this work, we propose a novel question-driven abstractive summarization method, Multi-hop Selective Generator (MSG), to incorporate multi-hop reasoning into question-driven summarization and, meanwhile, provide justifications for the generated summaries.
Specifically, we jointly model the relevance to the question and the interrelation among different sentences via a human-like multi-hop inference module, which captures important sentences for justifying the summarized answer. A gated selective pointer generator network with a multi-view coverage mechanism is designed to integrate diverse information from different perspectives. Experimental results show that the proposed method consistently outperforms state-of-the-art methods on two non-factoid QA datasets, namely WikiHow and PubMedQA. 
\end{abstract}

\section{Introduction}
\begin{figure*}
\centering
\includegraphics[width=\textwidth]{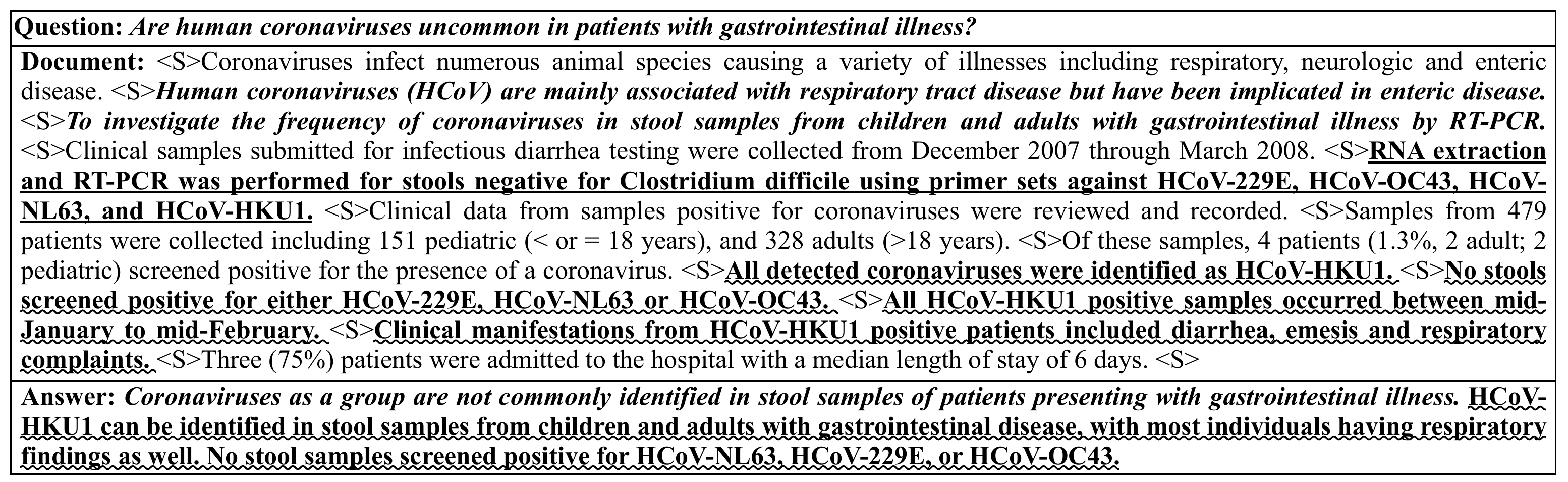}
\caption{An example from PubMedQA. The \textbf{highlighted} sentences illustrate the inference process when humans answer the given question. \textit{\textbf{Italic}} represents direct matching sentences from the question. \underline{\textbf{Underlined}} and \uwave{\textbf{wavy-underlined}} represent sentences inferred by 2nd-hop and 3rd-hop reasoning, respectively, to justify the answer.}
\label{example}
\vspace{-0.3cm}
\end{figure*}

Recent years have witnessed several attempts on exploring question-driven summarization, which aims at summarizing the source document with respect to a specific question, to produce a concise but informative answer in non-factoid question answering (QA)~\cite{DBLP:conf/acl/TomasoniH10,DBLP:conf/acl/ChanZWC12,DBLP:conf/wsdm/SongRLLMR17}. Unlike factoid QA~\cite{DBLP:conf/emnlp/RajpurkarZLL16}, e.g., ``Who is the author of Harry Potter?", whose answer is generally a single phrase or a short sentence with limited information, the answers for non-factoid questions are supposed to be more informative, involving some detailed analysis to explain or justify the final answers, such as questions in community QA~\cite{DBLP:conf/aaai/IshidaTOIKK18,DBLP:conf/aaai/DengLXCL0S20} or explainable QA~\cite{DBLP:conf/acl/FanJPGWA19,DBLP:conf/aaai/NakatsujiO20}. As the example from PubMedQA~\cite{DBLP:conf/emnlp/JinDLCL19} presented in Figure~\ref{example}, the answer can be regarded as the summary over the document driven by the reasoning process of the given question.

Most of related studies focus on query-based summarization approaches for summarizing the query-related content from the source document~\cite{DBLP:conf/icdm/ShenL11,DBLP:conf/acl/WangRCFC13,DBLP:conf/coling/CaoLLWL16,DBLP:conf/acl/NemaKLR17}. However, these approaches fall short of tackling question-driven summarization problem in QA scenario, since the query-based summarization process is typically based on semantic relevance measurement without a careful reasoning or inference process, which is essential to question-driven summarization. Currently, question-driven summarization is mainly explored by traditional information retrieval methods to select sentences from the source document to construct the final answer~\cite{DBLP:conf/coling/WangRCC14,DBLP:conf/wsdm/SongRLLMR17,DBLP:journals/tkde/YuliantiCSCS18}, which heavily rely on hand-crafted features or tedious multi-stage pipelines. Besides, compared to extractive summarization~\cite{DBLP:conf/coling/CaoLLWL16}, abstractive methods~\cite{DBLP:conf/acl/NemaKLR17} can produce more coherent and logical summaries to answer the given question.
To this end, we study question-driven abstractive summarization to generate natural form of answers by summarizing the source document with respect to a specific question.

To tackle question-driven abstractive summarization, the content selection process for summarization is not only determined by the semantic relevance to the given question, but it also requires a human-like reasoning and inference process to consider the content interrelationship comprehensively and carefully across the whole source text for generating the summary. For instance, in Figure~\ref{example}, given the specific question, there are several \textbf{highlighted} sentences required to be concentrated for conducting summarization so as to generate the answer. It leads to the necessity of measuring the importance of each sentence, instead of regarding the source text as an undifferentiated whole. 
Among these \textbf{highlighted} sentences, only the \textit{\textbf{italic}} sentences are directly related to the given question, while other \textbf{highlighted} sentences need to be inferred from their interrelationships with other sentences. In other words, the generated summary is likely to lose important information, if we only focus on the semantically relevant content to the given question. Moreover, it can be observed that one-time inference sometimes is insufficient for collecting all the required information for producing a summary. In this example, the answer is summarized from both the \textbf{\textit{1st-hop}} and \uwave{\textbf{3rd-hop}}  inference sentences in the document, indicating the importance of multi-hop reasoning for content selection in question-driven summarization.

In this work, we propose a question-driven abstractive summarization model, namely Multi-hop Selective Generator (MSG), which incorporates multi-hop inference to summarize abstractive answers over the source document for non-factoid questions. Concretely, the document is regarded as a hierarchical text structure to be assessed with the importance degree in both word- and sentence-level for content selection. Then we develop a multi-hop inference module to enable human-like multi-hop reasoning in question-driven summarization, which considers the semantic relevance to the question as well as the information consistency among different sentences. Finally, a gated selective pointer generator network with multi-view coverage mechanism is proposed to generate a concise but informative summary as the answer to the given question.

The main contributions of this paper can be summarized as follows: (1) We propose a novel question-driven abstractive summarization model for generating answers in non-factoid QA, which incorporates multi-hop reasoning to infer the important content for facilitating answer generation; (2) We propose a multi-view coverage mechanism to address the repetition issue along with the multi-view pointer network and generate informative answers; (3) Experimental results show that the proposed method achieves state-of-the-art performance on WikiHow and PubMedQA datasets, and it is able to provide justification sentences as the evidence for the answer.

\section{Related Works}
\noindent\textbf{Query-based Summarization.} 
Early works on query-based summarization focus on extracting query-related sentences to construct the summary~\cite{DBLP:conf/naacl/LinMD10,DBLP:conf/icdm/ShenL11}, which are later improved by exploiting sentence compression on the extracted sentences~\cite{DBLP:conf/acl/WangRCFC13,DBLP:conf/coling/LiL14}. Recently, some data-driven neural abstractive models are proposed to generate natural form of summaries with respect to the given query~\cite{DBLP:conf/acl/NemaKLR17,DBLP:journals/corr/abs-1712-06100}. However, current studies on query-based abstractive summarization are restricted by the lack of large-scale datasets~\cite{DBLP:conf/aaai/BaumelCE16,DBLP:conf/acl/NemaKLR17}. One the other hand, some researchers spark a new pave of question-driven summarization in non-factoid QA~\cite{DBLP:conf/wsdm/SongRLLMR17,DBLP:journals/tkde/YuliantiCSCS18,DBLP:conf/sigir/DengZL0LS20}, which requires the ability of reasoning or inference for supporting summarization, not merely relevance measurement, and also preserves remarkable testbeds of large-scale datasets.

\noindent\textbf{Non-factoid Question Answering.}
Different from factoid QA that can be tackled by extracting answer spans~\cite{DBLP:conf/emnlp/RajpurkarZLL16} or generating short sentences~\cite{DBLP:conf/nips/NguyenRSGTMD16,DBLP:journals/tacl/KociskySBDHMG18}, non-factoid QA aims at producing relatively informative and complete answers. In the past studies, non-factoid QA focused on retrieval-based methods, such as answer sentence selection~\cite{DBLP:conf/semeval/NakovMMMGR15} or answer ranking~\cite{DBLP:conf/sigir/ZhangDL20}. Recently, several efforts have been made on tackling long-answer generative question answering over supporting documents, which targets on questions that require detailed explanations \cite{DBLP:conf/acl/FanJPGWA19}. This kind of QA problem contains a large proportion of non-factoid questions, such as ``how" or ``why" type questions \cite{DBLP:journals/corr/abs-1810-09305,DBLP:conf/aaai/IshidaTOIKK18,DBLP:conf/aaai/DengLXCL0S20}. Besides, some studies aim at generating a conclusion for the concerned question
\cite{DBLP:conf/emnlp/JinDLCL19,DBLP:conf/aaai/NakatsujiO20}. \newcite{DBLP:conf/acl/FanJPGWA19} propose a multi-task Seq2Seq model with the concatenation of the question and support documents to generate long-form answers. \newcite{DBLP:conf/aaai/IidaKITOK19} and \newcite{DBLP:conf/aaai/NakatsujiO20} incorporate some background knowledge into Seq2Seq model for why questions and conclusion-centric questions. Some latest works~\cite{DBLP:conf/acl/FeldmanE19,DBLP:conf/emnlp/YadavBS19,DBLP:conf/acl/NishidaNNOSAT19} attempt to provide evidence or justifications for human-understandable explanation of the multi-hop inference process in factoid QA, where the inferred evidences are only treated as the middle steps for finding the answer. However, in non-factoid QA, the intermediate output is also important to form a complete answer, which requires a bridge between the multi-hop inference and summarization.

\section{Proposed Framework}
\begin{figure*}
\centering
\includegraphics[width=\textwidth]{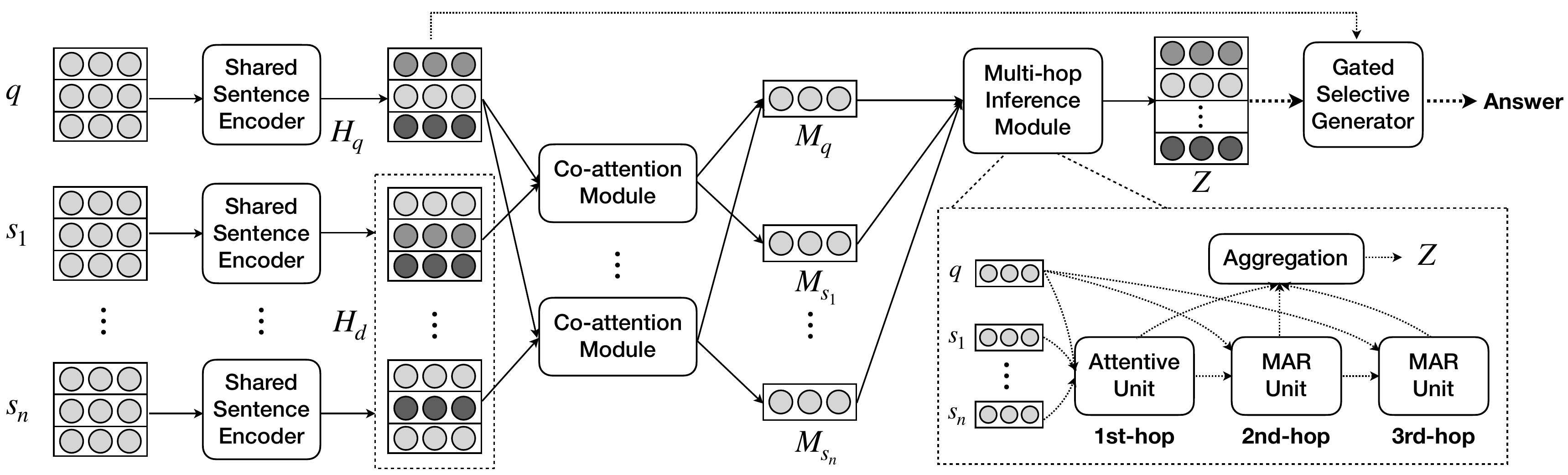}
\caption{The overview of Multi-hop Selective Generator (MSG).}
\label{encoder}
\vspace{-0.3cm}
\end{figure*}

We propose a question-driven abstractive summarization model, namely Multi-hop Selective Generator (MSG). The overview of MSG is depicted in Figure~\ref{encoder}, which consists of three main components: (1) \textit{Co-attentive Encoder} (Section 3.1), (2) \textit{Multi-hop Inference Module} (Section 3.2), and (3) \textit{Gated Selective Generator} (Section 3.3). Moreover, \textit{Multi-view Coverage Loss} is integrated to the overall training procedure (Section 3.4).

\subsection{Co-attentive Encoder}
Pre-trianed word embeddings, $E_q$ and $E_{s_i}$, of the question $q$ and each sentence $s_i$ in the document $D=\{s_1,s_2,...,s_n\}$ are input into the model. We first encode the question and each sentence in the document by a Bi-LSTM (Bidirectional Long Short-Term Memory Networks) shared encoder to learn the word-level contextual information, $H_q, H_{s_i}\in\mathbb{R}^{l\times d_h}$, where $l$ and $d_h$ denotes the sentence length and the dimension of the encoder output respectively. The overall word-level representations $H_d$ for the document is sequentially concatenated by $[H_{s_1},H_{s_2},...,H_{s_n}]$.

We compute the attention weights to align the word-level information between the question and the document sentences, and obtain the attention-weighted vectors of each word for both the question and the document sentences. For the question $q$ and the $i$-th sentence $s_i$ in the document $D$, we have:
\begin{gather}
O_{q{s_i}}=\text{tanh}\left(H_q^TUH_{s_i}\right),\\
\alpha_{q_i} = \text{softmax}(\text{Max}(O_{q{s_i}})),\\
\alpha_{s_i} =\text{softmax}(\text{Max}({O_{q{s_i}}}^T)),
\end{gather}
where $U\in\mathbb{R}^{d_{h}\times{d_{h}}}$ is the attention matrix to be learned; $\alpha_{q_i}$ and $\alpha_{s_i}$ are co-attention weights for the question and $i$-th sentence in the document.

We conduct dot product between the attention vectors and the word-level representations to generate the sentence representations for the question and the document:
\begin{gather}
 M_q = \frac{1}{n}\sum\nolimits_{i=1}^n H_q^T  \alpha_{q_i} \\
 M_s = [H_{s_1}^T \alpha_{s_1}:...:H_{s_n}^T  \alpha_{s_n}],
\end{gather}
where $M_q$ and $M_s$ denote the sentence-level representations for the question and the document.

\subsection{Multi-hop Inference Module}
Multi-hop Inference Module measures the degree of importance for each sentence in the document to generate the answer, through a multi-hop reasoning procedure, which contains two kinds of inference units: Attentive Unit and MAR Unit.

\subsubsection{Attentive Unit}
Attentive Unit basically measures the matching degree between each sentence in the document and the given question by the following vanilla attention mechanism:\vspace{-0.1cm}
\begin{gather}
    m_{dq} = \text{tanh}(M_s W_m M_q),\\
    \alpha_s = \text{softmax}(\omega_m^T m_{d_q}),\label{eq7}\\
    \mathbf{Attentive}(M_s,M_q) = M_s \odot \alpha_s,
\end{gather}
where $W_m$ and $\omega_m$ are the attention matrices to be learned. $\alpha_s$ is the sentence-level attention weight which measures the matching degree of each document sentence with the given question. $\odot$ denotes the element-wise product for obtaining the attentive sentence-level representations for the document. 

\subsubsection{MAR Unit}
Maximal Marginal Relevance (MMR) is an IR model that can be adopted to measure the query-relevancy and information-redundancy simultaneously for extractive summarization~\cite{DBLP:conf/sigir/CarbonellG98}. However, as for the content selection in abstractive summarization, the relevance to both the question and the other sentences in the document should be taken into consideration for a high recall of selecting necessary content. Thus, we propose Maximal Absolute Relevance (MAR) to select highly salient sentences for generating the summary, which is formulated as:
\begin{equation}\label{mar}
\begin{split}
mar_i = &\lambda\text{Sim}_1(M_{s_i},M_q)+\\
&(1-\lambda)\max_{s_j\in D,j\neq i}\text{Sim}_2(M_{s_i},M_{s_j}),
\end{split}
\end{equation}
where $\lambda$ is a hyper-parameter for balancing the question-relevancy and information-consistency measurement. The relevance to the question is calculated by:
\begin{equation}
\text{Sim}_1(M_{s_i},M_q) = M_{s_i}U_{1}M_q,
\end{equation}
where $U_{1}$ is a similarity matrix to be learned. We apply an attention mechanism over other sentences in the document to choose the highest relevance score, which can be regarded as the reasoning procedure where the next-hop justification sentences are supposed to be highly related to the last-hop justification sentences. 
\begin{gather}
e_{ij} = \tanh(M_{s_i}U_{2}M_{s_j}),\\
\text{Sim}_2(M_{s_i},M_{s_j}) = \frac{exp(e_{ij})}{\sum\nolimits_j exp(e_{ij})},
\end{gather}
where $U_{1}$ is a similarity matrix to be learned. 

Then the weighted sentence representations are computed by the element-wise product of the original sentence representations and the MAR scores gated by a sigmoid function denoted as $\sigma$:
\begin{gather}
%\hat{mar}_i = \frac{exp(mar_{i})}{\sum\nolimits_i exp(mar_i)},\\
\mathbf{MAR}(M_s,M_q) = M_s \odot \sigma(mar).
\end{gather}

Overall, MAR Unit assigns higher weights to sentences in two situations: (i) Those sentences are correlated to the given question, due to the first term in Equation~\ref{mar}, (ii) Those sentences are consistent with the highly weighted justification sentences from the last hop, due to the second term.

\subsubsection{Reasoning Procedure}
In accordance with human-like multi-hop inference procedure, the first hop is supposed to capture the semantic-relevant sentences to the given question. Then the subsequent hops should consider not only the relevance to the question, but also the information-consistency with the previous attended sentences. Hence, the Attentive Unit is adopted as the 1st-hop inference unit, while the MAR Unit is served as the $k$th-hop unit, where $k>1$. Before each hop, a Bi-LSTM layer is employed to refine the input sentence representation. For instance, a 3-hop inference procedure is as follows:
\begin{gather}
    M_s^{(1)} = \mathbf{Attentive}(\text{Bi-LSTM}(M_s), M_q),\\
    M_s^{(2)} = \mathbf{MAR}(\text{Bi-LSTM}(M_s^{(1)}), M_q),\\
    M_s^{(3)} = \mathbf{MAR}(\text{Bi-LSTM}(M_s^{(2)}), M_q).
\end{gather}

Then, we merge the 3-hop sentence representations, $\hat{M}_s = [M_s^{(1)},M_s^{(2)},M_s^{(3)}]$, via the following attention mechanism:
\begin{gather}
    \alpha_h = \text{softmax} (\omega_h^T \tanh(W_h\hat{M}_s)),\\
    Z = \hat{M}_s^T  \alpha_h,
\end{gather}
where $W_h$ and $\omega_h$ are attention matrices to be learned. $Z$ is the final sentence-level document representation for justifying the importance degree of each sentence in the decoding phase.

\begin{figure}
\setlength{\abovecaptionskip}{2pt}   
\setlength{\belowcaptionskip}{2pt}
\centering
\includegraphics[width=0.48\textwidth]{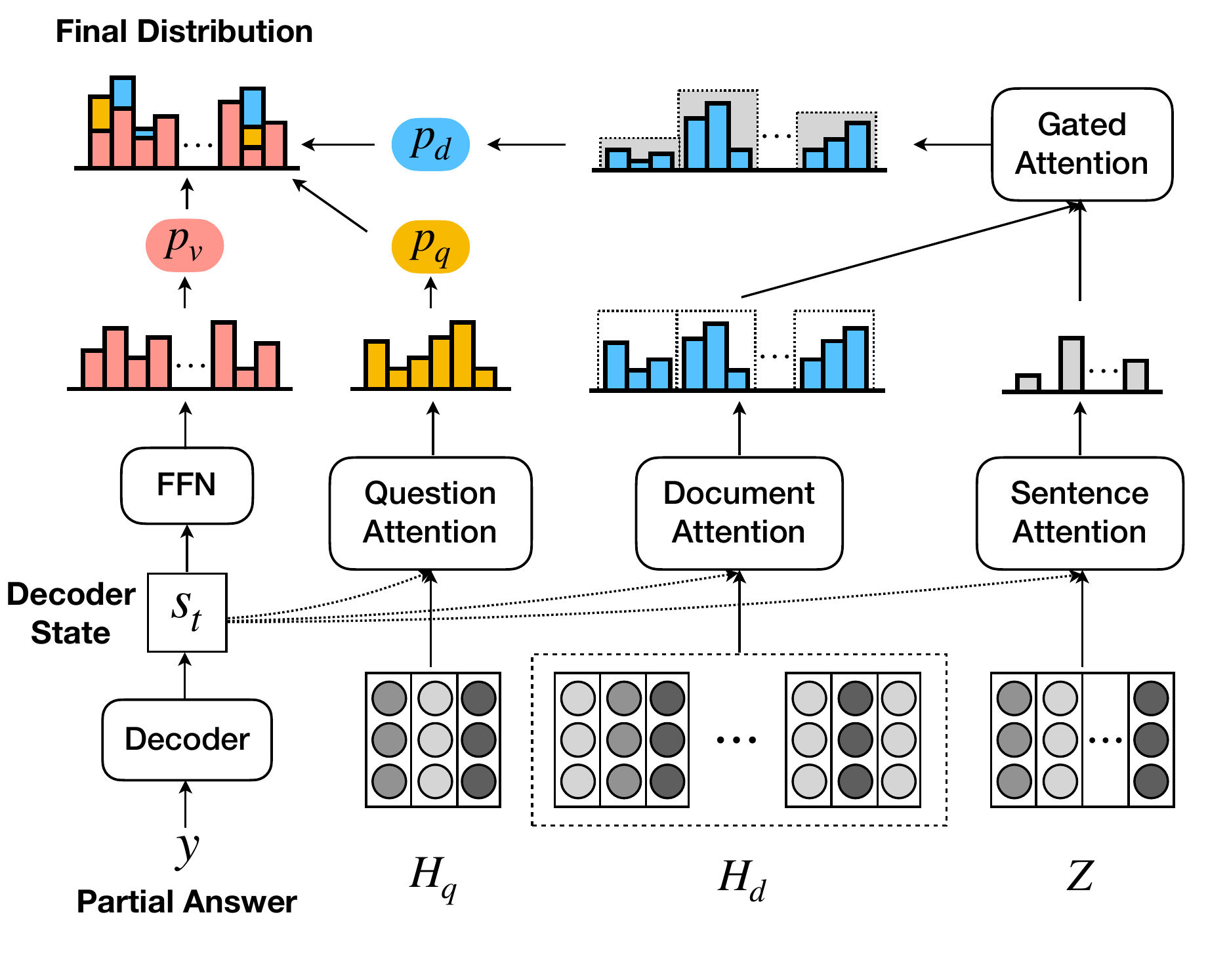}
\caption{Gated Selective Pointer-Generator Network.}
\label{decoder}
\vspace{-0.3cm}
\end{figure}

\subsection{Gated Selective Generator}
We obtain the word-level representations $H_q$ and $H_d$ for the question and document, respectively, from the encoding phase, and the sentence-level document  representation $Z$ via the multi-hop inference module. Figure~\ref{decoder} depicts the Gated Selective Pointer Generator Network in MSG.

%\subsubsection{Decoder}
A unidirectional LSTM is adopted as the decoder.  At each step $t$, the decoder produces hidden state $s_t$ with the input of the previous word $w_{t-1}$. 
The attention for each word in the question and the document, $\alpha^q_t$ and $\alpha^d_t$, are generated by:\vspace{-0.1cm}
\begin{gather}
e^{q_j}_t = {\omega^q_t}^T \text{tanh} (W_{q} H_{q_j}+W_{qs} s_t+b_{q}), \label{q_att}\\
\alpha^q_t = \text{softmax}(e^q_t),\\
e^{d_i}_t = {\omega^d_t}^T \text{tanh} (W_{d} H_{d_i}+W_{ds} s_t+b_{d}), \label{d_att}\\
\alpha^d_t = \text{softmax}(e^d_t),
\end{gather}
where $W_{q}$, $W_{qs}$, $W_{d}$, $W_{ds}$, $\omega^q_t$, $\omega^d_t$, $b_{q}$, $b_{d}$ are parameters to be learned. 

%\subsubsection{Gated Attention}
Then, we incorporate the multi-hop inference results $Z$ to compute the gated attention weights $\beta_t$ for each sentence in the document:
\begin{gather}
    \beta_t = \sigma({\omega^s_t}^T \text{tanh} (W_{s} Z_{k}+W_{ss} s_t+b_{s})),
\end{gather}
where $W_{s}$, $W_{ss}$, $\omega^s_t$, $b_{s}$ are parameters to be learned. 
We re-weight the word-level document attention scores $\alpha^d$ gated by the sentence-level document attention scores $\beta$ to attend important justification sentences along with the decoding process:
\begin{equation}
    \hat{\alpha}^{d_i}_t = \frac{\alpha^{d_i}_t \beta_{t, d_i \in s_k}}{\sum\nolimits_i \alpha^{d_i}_t  \beta_{t, d_i \in s_k}}.
\end{equation}
Thus, the re-weighted word-level document attention $\hat{\alpha}^{d}$ naturally blends with the results from the multi-hop inference module to enhance the influence of those important justification sentences.

%\subsubsection{Multi-view Pointer-Generator}
Finally, a multi-view pointer-generator architecture is designed to generate answers with multi-hop inference results as well as handle the multi-perspective out-of-vocabulary (OOV) issue. Such approach enables MSG to copy words from the question and be aware of the differential importance degree of different sentences in the document. 

The attention weights $\alpha^q_t$ and $\hat{\alpha}^d_t$ are used to compute context vectors $c^q_t$ and $c^d_t$ as the probability distribution over the source words:
\begin{gather}
    c^q_t =  H_{q}^T\alpha^{q}_t , \quad
    c^d_t =  H_{d}^T\hat{\alpha}^{d}_t .
\end{gather}

The context vector aggregates the information from the source text for the current step. We concatenate the context vector with the decoder state $s_t$ and pass through a linear layer to generate the answer representation $h^s_t$:
\begin{equation}
    h^s_t=W_1[s_t:c^q_t:c^d_t]+b_1,
\end{equation}
where $W_1$ and $b_1$ are parameters to be learned.

Then, the probability distribution $P^v$ over the fixed vocabulary is obtained by passing the answer representation $h^s_t$ through a softmax layer:
\begin{equation}
    P^v(y_t) = \text{softmax}(W_2 h^s_t + b_2),
\end{equation}
where $W_2$ and $b_2$ are parameters to be learned.

The final probability distribution of $y_t$ is obtained from three views of word distributions:
\begin{gather}
    P^q(y_t) = \sum\nolimits_{i:w_i=w} \alpha^{q_i}_t,\\
    P^d(y_t) = \sum\nolimits_{i:w_i=w} \hat{\alpha}^{d_i}_t,\\
    P^{all}(y_t) = [P^v(y_t),P^q(y_t),P^d(y_t)],\\
    \rho = \text{softmax}(W_{\rho}[s_t:c^q_t:c^d_t] + b_{\rho}),\\
    P(y_t) = \rho \cdot P^{all}(y_t),
\end{gather}
where $W_\rho$ and $b_\rho$ are parameters to be learned, $\rho$ is the multi-view pointer scalar to determine the weight of each view of the probability distribution. 

\subsection{End-to-end Training}
\textbf{Multi-view Coverage Loss.}
The original \textit{coverage mechanism}~\cite{DBLP:conf/acl/SeeLM17} could only prevent repeated attention from one certain source text. However, the repetition problem becomes more severe, as we leverage both the question and document as the source text. Besides, similar to multi-view pointer network, coverage losses of different sources are supposed to be weighted by their contribution. Therefore, we design a multi-view coverage mechanism to address this issue as well as balance the generating and copying processes. 

In each decoder timestep $t$, the coverage vector $c_t=\sum^{t-1}_{t'=0}a_{t'}$ is used to represent the degree of coverage so far. The coverage vector $c_t$ will be applied to compute the attention weight $\alpha_t$ in Equations~\ref{q_att} and~\ref{d_att}. 
The coverage loss is trained to penalize the repetition in updated attention weight $\alpha^t$ from all views. The re-normalized pointer weights $\hat{\rho} = \rho^c / \sum\nolimits_{c\in\{q,d\}} \rho^c$ are employed to balance the coverage loss of different views:
\begin{equation}
    L_{cov}=\sum\hat{\rho}\frac{1}{T}\sum\nolimits^T_{t=1}\sum\nolimits_i \text{min}(\alpha_t^i,c_t^i).
\end{equation}

\noindent \textbf{Overall Loss Function.}
The overall model is trained to minimize the negative log likelihood and the multi-view coverage loss:
\begin{equation}
    L=- \frac{1}{T}\sum\nolimits^T_{t=0}\text{log}P(w_t^*) + \lambda L_{cov},
\end{equation}
where $\lambda$ is a hyper-parameter to balance losses.

\section{Experiments}
\subsection{Datasets and Evaluation Metrics}
We evaluate on a large-scale summarization dataset with non-factoid questions, WikiHow~\cite{DBLP:journals/corr/abs-1810-09305}, and a non-factoid QA dataset with abstractive answers, PubMedQA~\cite{DBLP:conf/emnlp/JinDLCL19}. 
WikiHow is an abstractive summarization dataset collected from a community-based QA website, \textit{WikiHow}\footnote{https://www.wikihow.com}, in which each sample consists of a non-factoid question, a long article, and the corresponding summary as the answer to the given question. 
PubMedQA is a conclusion-based biomedical QA dataset collected from \textit{PubMed}\footnote{https://www.ncbi.nlm.nih.gov/pubmed/} abstracts, in which each instance is composed of a question, a context, and an abstractive answer which is the summarized conclusion of the context corresponding to the question. 
The statistics of the WikiHow and PubMedQA datasets are shown in Table~\ref{stat}~\footnote{https://github.com/dengyang17/msg}. We adopt ROUGE F1 (R1, R2, RL) for automatically evaluating the summarized answers. Besides, human evaluation and Distinct scores are adopted for analysis. 

\begin{table}
\fontsize{9}{10}\selectfont
\centering
  \begin{tabular}{ccc}
  \toprule
  Dataset  & \multirow{2}{*}{WikiHow}&\multirow{2}{*}{PubMedQA}\\
  (train/dev/test)&&\\
  \midrule
  \#Samples & 168K / 6K / 6K  & 169K / 21K / 21K  \\
  Avg QLen &7.00 / 7.02 / 7.01 & 16.3 / 16.4 / 16.3 \\
  Avg DLen &582 / 580  / 584 & 238 / 238 / 239  \\
  Avg ALen &62.2 / 62.2 / 62.2 & 41.0 / 41.0 / 40.9 \\
  Avg \#Sents/Doc & 20.7 / 20.7 / 20.6 & 9.32 / 9.31 / 9.33 \\
  \bottomrule
  \end{tabular}
  \caption{Statictis of Dataset}
\label{stat}
\vspace{-0.3cm}
\end{table}

\subsection{Baseline Methods and Implementations}
To evaluate the proposed method, we compare with several baselines and state-of-the-art methods on query-based abstractive summarization and generative QA. We first employ four widely-adopted summarization baseline methods, including two unsupervised extractive methods, \textbf{LEAD3} and \textbf{MMR}, and two abstractive methods, \textbf{S2SA}~\cite{DBLP:journals/corr/BahdanauCB14}, and \textbf{PGN}~\cite{DBLP:conf/acl/SeeLM17}. 

Then two popular query-based abstractive summarization methods are adopted for evaluation: (1) \textbf{SD$_2$}~\cite{DBLP:conf/acl/NemaKLR17}, which is a sequence-to-sequence model with a query attention, and (2) \textbf{QS}~\cite{DBLP:journals/corr/abs-1712-06100}, which incorporates question information into the pointer-generator network with the vanilla attention mechanism. 

Finally, we implement two latest generative QA models for comparisons: (1)  \textbf{S2S-MT}~\cite{DBLP:conf/acl/FanJPGWA19}, which uses a multi-task Seq2Seq model with the concatenation of question and support document, and (2) \textbf{QPGN}~\cite{DBLP:conf/aaai/DengLXCL0S20}, which is a question-driven pointer-generator network with co-attention between the question and document.

We train all the models with pre-trained GloVE embeddings\footnote{http://nlp.stanford.edu/data/glove.42B.zip} of 300 dimensions and set the vocabulary size to 50k. During training and testing procedure, we restrict the length of generated summaries within 50 words. As for the proposed method, we train with a learning rate of 0.15 and an initial accumulator value of 0.1. The dropout rate is set to 0.5. The hidden unit sizes of the BiLSTM encoder and the LSTM decoder are all set to 256. We train our models with the batch size of 32. All other parameters are randomly initialized from [-0.05, 0.05]. Similar to the original coverage loss~\cite{DBLP:conf/acl/SeeLM17}, we first train the model without multi-view coverage loss for 20 epochs, and then train with it for another 5 epochs with $\lambda$ as 0.1. 

\subsection{Performance Comparison}\label{sec:results}

\begin{table}
\fontsize{10}{11}\selectfont
\centering
\setlength{\tabcolsep}{1.3mm}{
\begin{tabular}{lcccccc}
\toprule
\multirow{2}{*}{Model}& \multicolumn{3}{c}{WikiHow}& \multicolumn{3}{c}{PubMedQA}\\ 
\cmidrule(lr){2-4}\cmidrule(lr){5-7}
&R1&R2&RL&R1&R2&RL\\
\midrule
\midrule
LEAD3&26.0$^*$& 7.2$^*$& 24.3$^*$&30.9&9.8&21.2\\
MMR&26.8&6.1&23.6&30.1&9.0&24.4\\
\midrule
S2SA&22.0$^*$& 6.3$^*$& 20.9$^*$ &32.4&11.0&27.3\\
PGN&28.5$^*$& 9.2$^*$& 26.5$^*$ &32.9&11.5&28.1 \\
%TAG& 28.4$^\dagger$& 9.1$^\dagger$& 27.5$^\dagger$&-&-&- \\ 
\midrule
SD$_2$&27.7& 7.9& 25.8&32.3& 10.5&26.0\\
QS &\underline{28.8} & \underline{9.9}& 27.6&32.6&11.1&26.7\\
%MHPGM &28.6&9.6&27.5&34.0&12.5&28.4\\
S2S-MT &28.6&9.6&27.5&33.2&12.2&27.8\\
QPGN &\underline{28.8}&9.7&\underline{27.7}&\underline{34.2}&\underline{12.8}&\underline{28.7}\\
\midrule
\midrule
MSG (1-Hop)&30.0&10.2&29.0&36.5&14.4&30.0\\
MSG (2-Hop)&30.2&10.3&29.1&37.0&14.7&\textbf{30.4}\\
MSG (3-Hop)&\textbf{30.5}&\textbf{10.5}&\textbf{29.3}&\textbf{37.2}&\textbf{14.8}&30.2\\
\bottomrule
\end{tabular}}
\caption{\label{result} Results on WikiHow and PubMedQA. $^*$ represents results reported from \newcite{DBLP:journals/corr/abs-1810-09305}.}
\end{table}

Table~\ref{result} summarizes the experimental results on both datasets. As for WikiHow, which is an abstractive summarization dataset with non-factoid questions, current query-based summarization (SD$_2$, QS) and generative QA approaches (S2S-MT, QPGN) barely improve the performance from traditional summarization approaches. It indicates that the question information is not fully exploited for summarization, while MSG outperforms all these methods with a noticeable margin, about 2\%.

Besides, since PubMedQA is a QA dataset with abstractive answers, we can observe that QPGN, which employs special design for modeling the interaction between the question and document, achieves relatively better performance than other summarization methods. Favorably MSG raises the state-of-the-art result by about 3\%. Furthermore, MSG achieves promising improvements via the multi-hop inference on these two datasets.

\begin{table}
\fontsize{10}{11}\selectfont
\centering
  \begin{tabular}{ccccc}
  \toprule
  Model & Info & Conc & Read & Corr\\
  \midrule
  SD$_2$ & 3.48&3.34&3.30&3.04\\
  QS &3.62&3.30&3.48&3.24 \\
  QPGN &3.58&3.52&3.68&3.32\\
  %\textbf{MSG (1-Hop)} & 4.08&4.20&4.22&3.58\\
  %\textbf{MSG (2-Hop)} & 4.06&4.32&4.36&3.60\\
  \textbf{MSG} & \textbf{4.14}&\textbf{3.88}&\textbf{3.82}&\textbf{3.78}\\
  \bottomrule
  \end{tabular}
\caption{Human Evaluation Results}
\label{human_eval}
\vspace{-0.4cm}
\end{table}

%\subsection{Human Evaluation}
We conduct human evaluation to evaluate the generated answer from four aspects: (1) Informativity: how rich is the generated answer in information? (2) Conciseness: how concise is the summary? (3) Readability: how fluent and coherent is the summary?  (4) Correctness: how well does the generated answer respond to the given question?  We randomly sample 50 questions from two datasets and generate their answers with three query-based summarization methods, including SD$_2$, QS, QPGN and the proposed MSG. Three annotators are asked to score each generated answer with 1 to 5 (higher the better). Results are presented in Table~\ref{human_eval}. We observe that MSG consistently and substantially outperforms existing query-based summarization methods in all aspects, especially for the informativeness and correctness. The results show that MSG effectively generates concise but also informative answers, since MSG not only considers question-related information, but also captures logically necessary content for answering the given question via multi-hop reasoning. Consequently, it leads to a more precise answer.

\section{Discussions}
\subsection{Ablation Study}
\begin{table}
\fontsize{10}{11}\selectfont
\centering
  \begin{tabular}{lcccc}
    \toprule
    \multirow{2}{*}{Model}& \multicolumn{2}{c}{WikiHow}& \multicolumn{2}{c}{PubMedQA}\\ 
    \cmidrule(lr){2-3}\cmidrule(lr){4-5}  
    &R1&RL&R1&RL\\
    \midrule
    \textbf{MSG (3-Hop)}&30.5&29.3&37.2&30.2\\
    \midrule
    - multi-hop inference & 29.5&28.4&35.7&29.2 \\
    ~~ - hops aggregation$^1$&30.1&29.0&37.0&30.1\\
    ~~ - hops attention&30.3&29.2&37.0&30.1\\
    ~~ - MAR unit$^2$&30.0&29.1&36.8&30.0\\
    %\textbf{MSG (1-Hop)}&30.0&29.0&36.5&30.0\\
    \midrule
    - co-attention&30.2&29.0&37.0&30.1\\
    %- sentence selector&29.5&28.4&35.7&29.2\\
    - gated attention$^3$&30.2&28.9&36.6&29.8\\
    - question pointer&30.3&29.1&35.5&29.1\\
    - MVC loss& 29.6&28.5&35.9&29.3\\
  \bottomrule
\end{tabular}
\caption{Ablation Study on Model Components. $^1$Use the sentence representation learned from the last hop, instead of merging all the hops. $^2$Replace all the MAR Unit with Attentive Unit. $^3$Replace the \textit{sigmoid} function with \textit{softmax} function.}
\label{ablation1}
\vspace{-0.4cm}
\end{table}

We conduct ablation study to validate the effectiveness of different components in MSG as well as the detailed design for the multi-hop inference module. 
The upper part in Table~\ref{ablation1} presents the ablation study on multi-hop inference module. First of all, the model performance suffers a great decrease from discarding the multi-hop inference module on two datasets, showing the necessity of incorporating the multi-hop reasoning into the question-driven summarization. In specific, the fusion of the selective sentence representations from all hops brings performance improvement, including aggregating all the hops as well as applying attention to weight the importance of each hop. Besides, it also achieves better performance to apply the proposed MAR Unit as the multi-hop unit, instead of repeatedly using Attentive Unit, indicating that it is not enough to only consider the question-related information, while the interrelationship among different sentences also attaches great importance.

The second part in Table~\ref{ablation1} presents the ablation study in terms of discarding other model components in MSG. In general, all the components contribute to the final performance to a certain extent. In detail, there are several notable observations: 
(1) Some existing works~\cite{DBLP:conf/acl/SunHLLMT18,DBLP:conf/acl/NishidaSNSOAT19} apply softmax function to normalize the weights of different sentences in the decoding phase, which falls short of differentiating the importance degree of each sentence. The result shows that MSG achieves better performance by employing gated attention to distinguish salient justification sentences for generating the summaries. (2) Discarding the question pointer casts a noticeably greater decrease on PubMedQA than WikiHow. We conjecture that those questions from PubMedQA contain more words available to be copied for generating precise summaries, as the statistic of the question length shown in Table~\ref{stat}. These results also validate the importance of multi-view PGN on question-driven abstractive summarization, which is underutilized in current methods. (3) Multi-view coverage (MVC) loss makes a great contribution to the performance by alleviating the severe repetition problem along with the multi-view PGN.

\subsection{Analysis of Multi-hop Reasoning}
\begin{figure}
\centering
\includegraphics[width=0.48\textwidth]{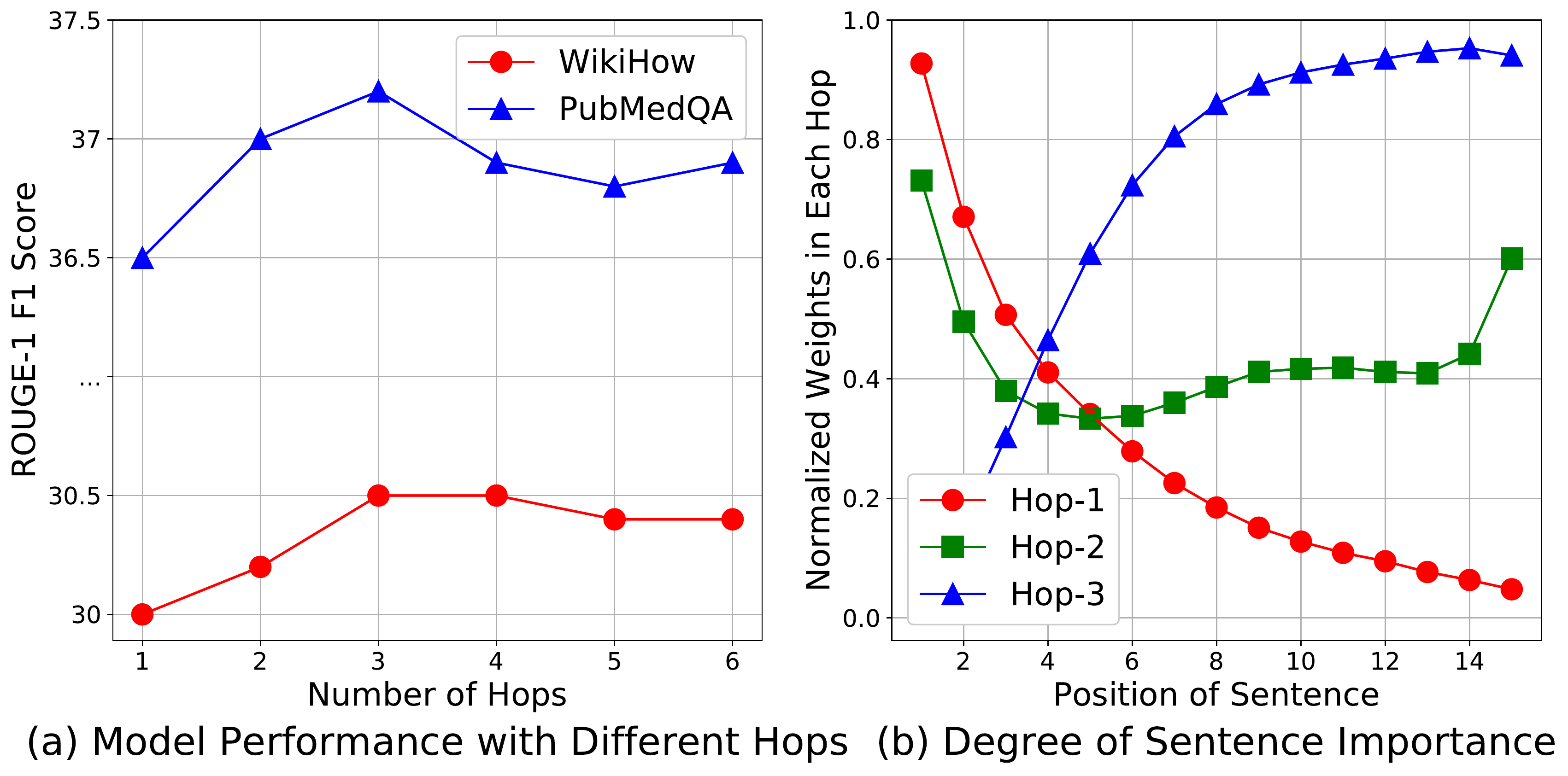}
\caption{Analysis of Multi-hop Reasoning}
\label{hops}
\vspace{-0.4cm}
\end{figure}
As the results presented in Section~\ref{sec:results}, MSG (3-Hop) outperforms MSG (1-Hop) by 0.5\% and 0.7\% on WikiHow and PubMedQA, respectively, indicating the effectiveness of incorporating multi-hop reasoning in question-driven summarization. Figure~\ref{hops}(a) presents the model performance in terms of using different hops of reasoning. We can see that, as expected, the performance of the model begins with growth when increasing the number of hops for reasoning. However, the performance becomes generally unchanged (e.g., WikiHow) or even slightly decreases (e.g., PubMedQA) when we further increase the number of hops. In practice, it is actually unnecessary to reason for too many hops, which may cause over-fitting. And adopting 3-hops in the implementation can be regarded as a hyper-parameter that is tuned on the datasets.

In addition, we extract and normalize the sentence weights from Eq.~\ref{eq7}\&\ref{mar} to analyze some characteristics of the justification sentences in multi-hop inference. Figure~\ref{hops}(b) summarizes the statistic result of the sentence importance degree in each hop. We observe that the most important sentences in the 1st-hop of reasoning are likely to appear at the beginning of the document, while those in the 3rd-hop are concentrated in the latter part of the document. Comparatively, the important sentences in the 2nd-hop appear equally in all positions of the document. The results show that the proposed multi-hop inference procedure of justification sentences is generally in accordance with human-like reading habits.

\begin{figure*}
\centering
\includegraphics[width=\textwidth]{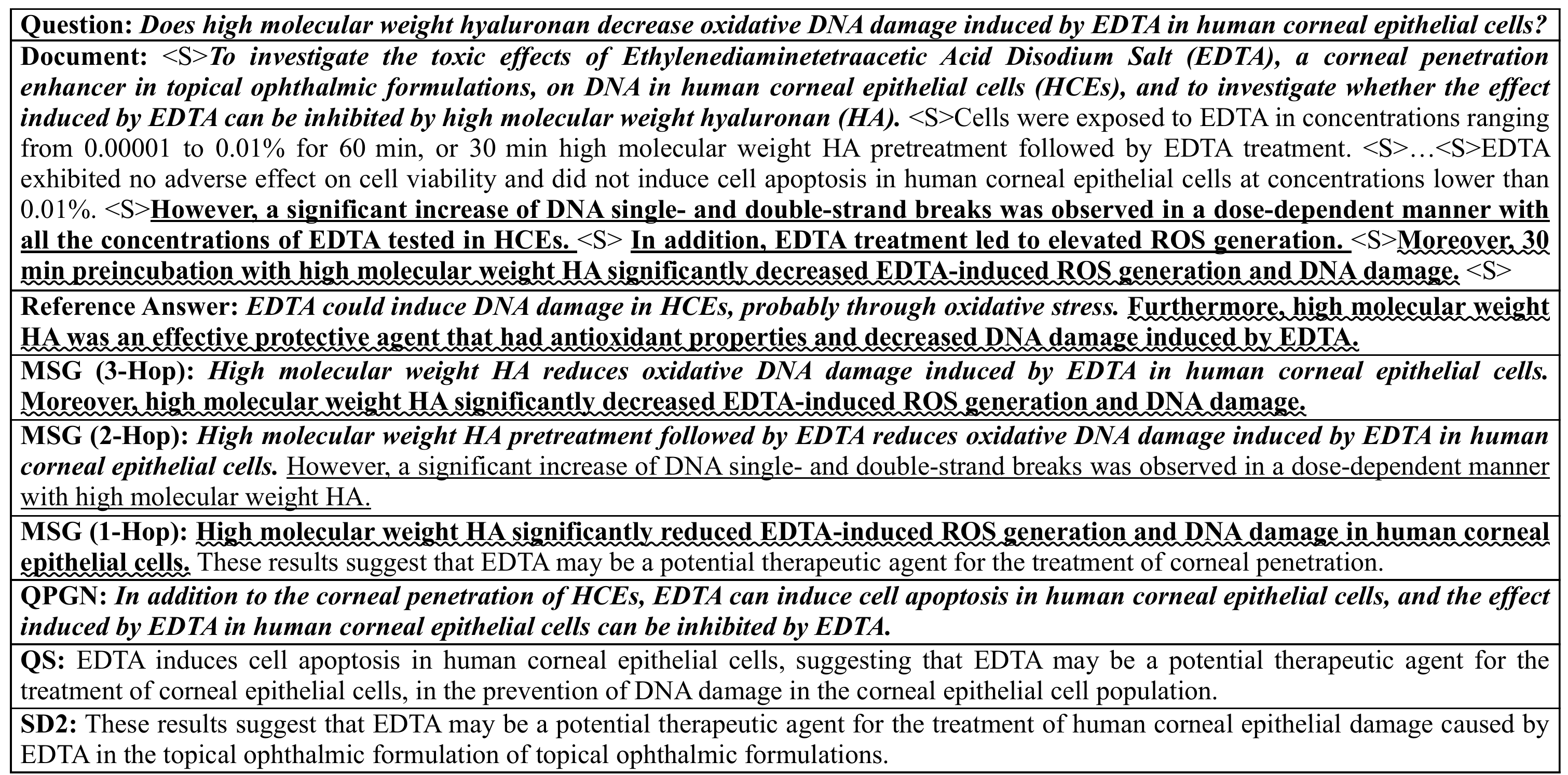}
\caption{A case study with the same legend as Figure~\ref{example}. The \textbf{highlighted} sentences are attended by MSG (3-hop).}
\vspace{-0.5cm}
\label{case12}
\end{figure*}
\subsection{Case Study}
We present a case study in Figure~\ref{case12} with generated answers from the proposed method and some baseline methods, QPGN, QS, and SD$_2$, to intuitively compare these methods.  With the multi-hop reasoning process in MSG, we can obtain a clear clue of how to answer the given question. As it can be observed that the reference answer is composed of the information from the \textbf{\textit{1st-hop}} and \textbf{\uwave{3rd-hop}} inference sentences, it is inadequate to simply summarize the question-related content for generating the answer. For the generated summaries, there are several observations as follows: (1) MSG (3-hop) successfully summarizes the source document with all the necessary and correct information. (2) MSG (2-hop) also effectively summarizes the 1st-hop and 2nd-hop inference content in the document. However, in this case, 3-hop inference is required to answer the given question. (3) MSG (1-hop) only measures the semantic relevance to the given question, leading to an incomplete summary that is lack of some necessary content, and even introduces some general sentences due to the data-driven learning. (4) QPGN only considers the semantic relevance to the given question, leading to an incomplete summary that is lack of some necessary content. (5) QS and SD$_2$ fail to capture the key information, 
resulting in generating irrelevant summaries to the given question, or producing some general sentences due to the data-driven learning. It shows the capability of MSG to implement multi-hop reasoning and provide justification sentences. 

Additionally, we observe that many cases probably require more than 3-hop inference or only involve one or two hops. However, we can still evaluate how MSG works in these cases. Compared to the reference answer, MSG (3-hop) can still capture most of the useful information to generate a good summary for answering the question. Besides, MSG (2-hop) and MSG (1-hop) also manage to attend some important content in the document. In general, our model is able to only attend a single hop if one-hop is enough, while our model may regard several hops as an integral hop when more hops are required. However, the baseline methods introduce much unnecessary or even incorrect information into the summarized answers.

\subsection{Duplication Analysis in Answers}
\begin{figure}
\centering
\includegraphics[width=0.48\textwidth]{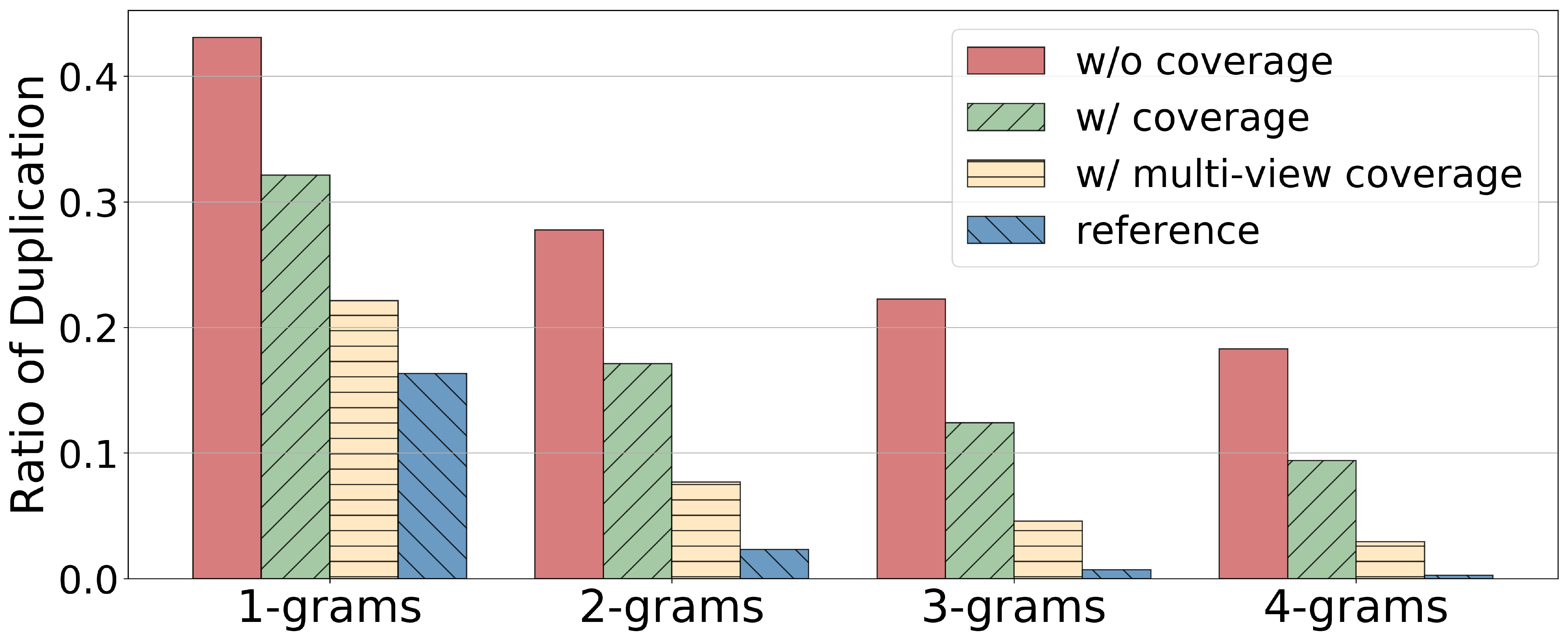}
\caption{Duplication Analysis in Answers}
\label{dup}
\vspace{-0.2cm}
\end{figure}

We adopt Distinct scores to analyze whether the multi-view coverage mechanism can alleviate the repetition issue in the generation procedure of multi-view PGN. Figure~\ref{dup} summarizes the percentage of n-grams duplication on the ground-truth answers and the generated answers with or without the original~\cite{DBLP:conf/acl/SeeLM17} and multi-view coverage mechanism. We observe that the original coverage mechanism can still reduce word repetition in multi-view PGN. Moreover, multi-view coverage further reduces the ratio of duplication to a great extent, since multi-view coverage not only prevents repeatedly attending to the same element in both question and document, but also balances the weight of penalty between them.

\section{Conclusion}
We propose a novel question-driven abstractive summarization method, Multi-hop Selective Generator (MSG), to summarize concise but informative answers for non-factoid QA. We incorporate multi-hop reasoning to infer justification sentences for abstractive summarization. Experimental results show that the proposed method achieves state-of-the-art performance on two benchmark non-factoid QA datasets, namely WikiHow and PubMedQA.

\bibliographystyle{acl_natbib}
\bibliography{emnlp2020.bib}

\iffalse
\begin{figure*}[htb]
%\setlength{\abovecaptionskip}{2pt}
%\setlength{\belowcaptionskip}{0pt}
\centering
\includegraphics[width=\textwidth]{lfqa-case3.pdf}
\caption{Case Study from WikiHow}
\label{case2}
\end{figure*}
\fi

\end{document}